\documentclass[10pt,twocolumn,letterpaper]{article}

\usepackage{iccv}
\usepackage{times}
\usepackage{epsfig}
\usepackage{graphicx}
\usepackage{amsmath}
\usepackage{amssymb}
 \usepackage{color}

\usepackage[pagebackref=true,breaklinks=true,letterpaper=true,colorlinks,bookmarks=false]{hyperref}

\usepackage[percent]{overpic}

\iccvfinalcopy 

\def\iccvPaperID{836} 

\ificcvfinal\fi
\begin{document}

\title{Active Object Localization with Deep Reinforcement Learning}

\author{Juan C. Caicedo\\
Fundaci\'on Universitaria Konrad Lorenz\\
Bogot\'a, Colombia\\
{\tt\small juanc.caicedor@konradlorenz.edu.co}
\and
Svetlana Lazebnik\\
University of Illinois at Urbana Champaign\\
Urbana, IL, USA\\
{\tt\small slazebni@illinois.edu}
}

\maketitle

\begin{abstract}
We present an active detection model for localizing objects in scenes. The model is class-specific and allows an agent to focus attention on candidate regions for identifying the correct location of a target object. This agent learns to deform a bounding box using simple transformation actions, with the goal of determining the most specific location of target objects following top-down reasoning. The proposed localization agent is trained using deep reinforcement learning, and evaluated on the Pascal VOC 2007 dataset. We show that agents guided by the proposed model are able to localize a single instance of an object after analyzing only between 11 and 25 regions in an image, and obtain the best detection results among systems that do not use object proposals for object localization.
\end{abstract}

\section{Introduction}
\label{sec:intro}

The process of localizing objects with bounding boxes can be seen as a control problem with a sequence of steps to refine the geometry of the box. Determining the exact location of a target object in a scene requires active engagement to understand the context, change the fixation point, identify distinctive parts that support recognition, and determine the correct proportions of the box.

During the last decade, the problem of object detection or localization has been studied by the vision community with the goal of recognizing the category of an object, and identifying its spatial extent with a tight bounding box that covers all its visible parts \cite{everingham2010pascal,russakovsky2014imagenet}. This is a challenging setup that requires computation and analysis in multiple image regions, and a good example of a task driven by active attention.

Important progress for improving the accuracy of object detectors has been recently possible with Convolutional Neural Networks (CNNs), which leverage big visual data and deep learning for image categorization. A successful model is the R-CNN detector proposed by Girshick et al. \cite{girshick2014rich, ren2015faster}, which combines object proposals and CNN features to achieve state-of-the-art results in the Pascal and ImageNet benchmarks. Several other works have proposed the use of high-capacity CNNs to directly predict bounding boxes under a regression setting also with good results \cite{sermanet2013overfeat, simonyan2014very, erhan2014scalable}.

\begin{figure}[t]
\begin{center}
  \includegraphics[width=1.00\linewidth]{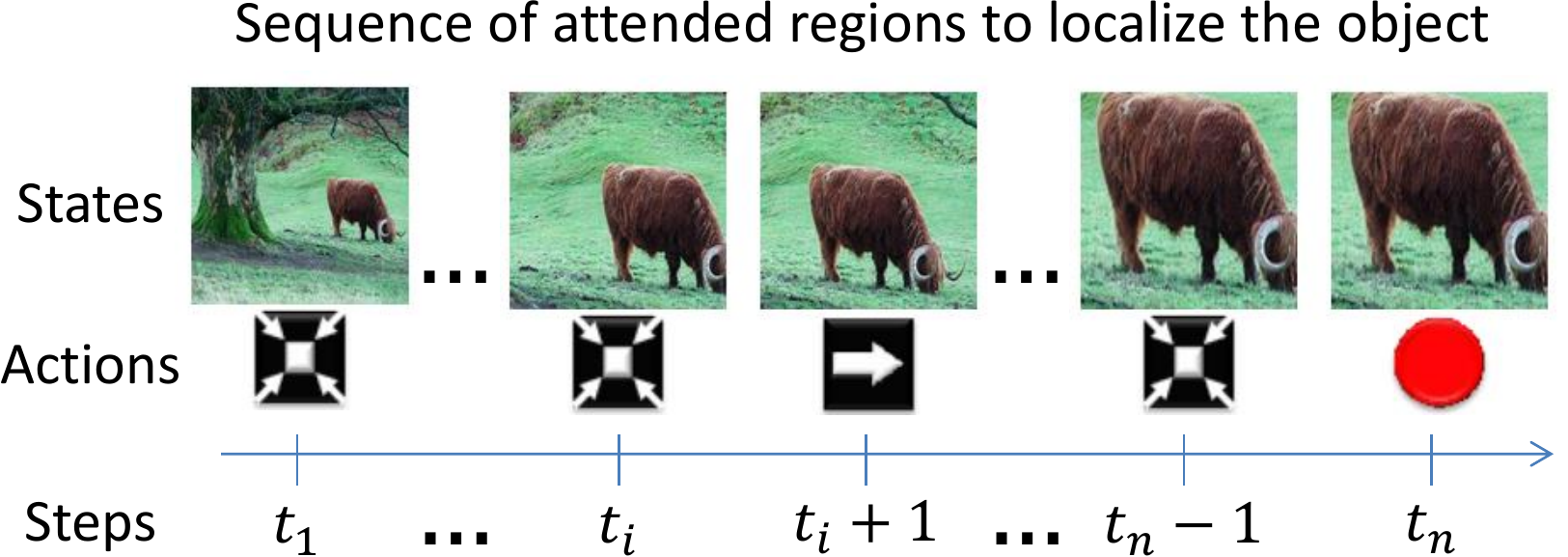}
\end{center}
   \caption{A sequence of actions taken by the proposed algorithm to localize a cow. The algorithm attends regions and decides how to transform the bounding box to progressively localize the object.}
\label{fig:intro}
\end{figure}

In this work, we propose a class-specific active detection model that \emph{learns} to localize target objects known by the system. The proposed model follows a top-down search strategy, which starts by analyzing the whole scene and then proceeds to narrow down the correct location of objects. This is achieved by applying a sequence of transformations to a box that initially covers a large region of the image and is finally reduced to a tight bounding box. The sequence of transformations is decided by an \emph{agent} that analyzes the content of the currently visible region to select the next best action. Each transformation should keep the object inside the visible region while cutting off as much background as possible. Figure \ref{fig:intro} illustrates some steps of the dynamic decision process to localize a cow in an image.

The proposed approach is fundamentally different from most localization strategies. In contrast to sliding windows, our approach does not follow a fixed path to search objects; instead, different objects in different scenes will end up in different search paths. Unlike object proposal algorithms, candidate regions in our approach are selected by a high-level reasoning strategy instead of following low-level cues. Also, compared to bounding box regression algorithms, our approach does not localize objects following a single, structured prediction method. We propose a dynamic \emph{attention-action} strategy that requires to pay attention to the contents of the current region, and to transform the box in such a way that the target object is progressively more focused.

To stimulate the attention of the proposed agent, we use a reward function proportional to how well the current box covers the target object. We incorporate the reward function in a reinforcement learning setting to learn a \emph{localization policy}, based on the DeepQNetwork algorithm \cite{mnih2015human}. As a result, the trained agent can localize a single instance of an object in about 11 steps, which means that the algorithm can correctly find an object after processing only 11 regions of the image. We conducted a comprehensive experimental evaluation in the challenging Pascal VOC dataset, obtaining competitive results in terms of precision and recall. In what follows, we present and discuss the components of the proposed approach and provide a detailed analysis of experimental results.

\section{Previous Works}
\label{sec:prevWorks}

Object localization has been successfully approached with sliding window classifiers. A popular sliding window method, based on HOG templates and SVM classifiers, has been extensively used to localize objects \cite{felzenszwalb2010object,malisiewicz2011ensemble}, parts of objects \cite{endres2013learning,lim2013parsing}, discriminative patches \cite{singh2012unsupervised,juneja2013blocks} and even salient components of scenes \cite{pandey2011scene}. Sliding windows are related to our work because they are category-specific localization algorithms, which is also part of our design. However, unlike our work, sliding windows make an exhaustive search over the location-scale space.

A recent trend for object localization is the generation of category independent object proposals. Hosang et al. \cite{hosang2014good} provide an in depth analysis of ten object proposal methods, whose goal is to generate the smallest set of candidate regions with the highest possible recall. Substantial acceleration is achieved by reducing the set of candidates in this way, compared to sliding windows. Nonetheless, object detection based on proposals follows the same design of window-based classification on a set of reduced regions, which is still large (thousands of windows) for a single image that may contain a few interesting objects.

Several works attempt to reduce the number of evaluated regions during the detection process. For instance, Lampert et al. \cite{lampert2008beyond} proposed a branch-and-bound algorithm to find high-scoring regions only evaluating a few locations. Recently, Gonzalez-Garcia et al. \cite{gonzalez2015active} proposed an active search strategy to accelerate category-specific R-CNN detectors. These methods are related to ours because they try to optimize computational resources for localization. Also related is the work of Divvala et al. \cite{divvala2009empirical}, which uses context to determine the localization of objects.

Visual attention models have been investigated with the goal of predicting where an observer is likely to orient the gaze (see \cite{borji2013state} for a recent survey). These models are generally based on a saliency map that aggregates low-level features to identify interesting regions. These models are designed to predict human fixations and evaluate performance with user studies \cite{torralba2006contextual}, while our work aims to localize objects and we evaluate performance in this task.

There is recent interest in attention capabilities for visual recognition in the machine learning community. Xu et al. \cite{xu2015show} use a Recurrent Neural Network (RNN) to generate captions for images, using an attention mechanism that explains where the system focused attention to predict words. Mnih et al. \cite{mnih2014recurrent} and Ba et al. \cite{ba2014multiple} also used RNNs to select a sequence of regions that need more attention, which are processed at higher resolution for recognizing multiple characters. Interestingly, these models are trained with Reinforcement Learning as we do; however, our work uses a simpler architecture and intuitive actions to transform boxes.

\section{Object Localization as a Dynamic Decision Process}
\label{sec:method}

We cast the problem of object localization as a Markov decision process (MDP) since this setting provides a formal framework to model an agent that makes a sequence of decisions. Our formulation considers a single image as the environment, in which the agent transforms a bounding box using a set of actions. The goal of the agent is to land a tight box in a target object that can be observed in the environment. The agent also has a state representation with information of the currently visible region and past actions, and receives positive and negative rewards for each decision made during the training phase. During testing, the agent does not receive rewards and does not update the model either, it just follows the learned policy.

Formally, the MDP has a set of actions $A$, a set of states $S$, and a reward function $R$. This section presents details of these three components, and the next section presents technical details of the training and testing strategies.

\subsection{Localization Actions}
\label{subsec:actions}

\begin{figure}[t]
\begin{center}
  \includegraphics[width=1.00\linewidth]{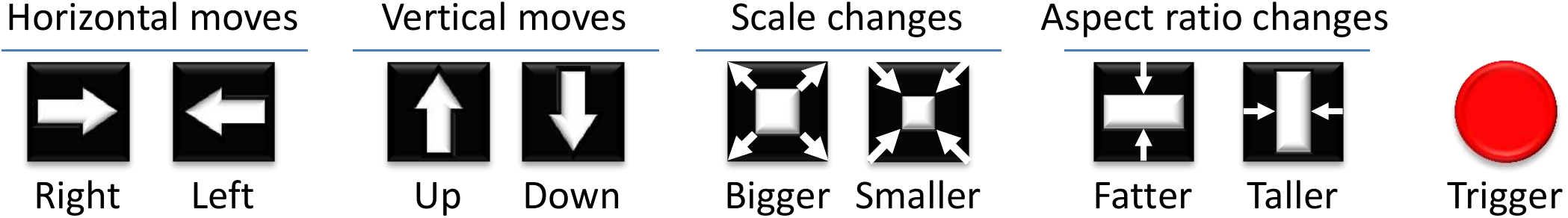}
\end{center}
   \caption{Illustration of the actions in the proposed MDP, giving 4 degrees of freedom to the agent for transforming boxes.}
\label{fig:actions}
\end{figure}

The set of actions $A$ is composed of eight transformations that can be applied to the box and one action to terminate the search process. These actions are illustrated in Figure \ref{fig:actions}, and are organized in four sub-sets: actions to move the box in the horizontal and vertical axes, actions to change scale, and actions to modify aspect ratio. In this way, the agent has four degrees of freedom to transform the box during any interaction with the environment.

A box is represented by the coordinates in pixels of its two corners: $b=[x_1, y_1, x_2, y_2]$. Any of the transformation actions make a discrete change to the box by a factor relative to its current size in the following way:
\begin{equation} \label{eq:changeFactor}
\alpha_w = \alpha*(x_2 - x_1) ~~~~~~ \alpha_h = \alpha*(y_2 - y_1)
\end{equation}
where $\alpha \in [0,1]$. Then, the transformations are obtained by adding or removing $\alpha_w$ or $\alpha_h$ to the $x$ or $y$ coordinates, depending on the desired effect. For instance, a horizontal move to the right adds $\alpha_w$ to $x_1$ and $x_y$, while decreasing aspect ratio subtracts $\alpha_w$ from $x_1$, and adds it to $x_2$. Note that the origin in the image plane is located in the top-left corner.

We set $\alpha=0.2$ in all our experiments, since this value gives a good trade-off between speed and localization accuracy. In early exploration experiments we noticed that smaller values make the agent slower to localize objects, while larger values make it harder to place the box correctly.

Finally, the only action that does not transform the box is a trigger to indicate that an object is correctly localized by the current box. This action terminates the sequence of the current search, and restarts the box in an initial position to begin the search for a new object. The trigger also modifies the environment: it marks the region covered by the box with a black cross as shown in the final frame of the top two examples in Figure \ref{fig:sequences}. This mark serves as a inhibition-of-return (IoR) mechanism by which the currently attended region is prevented from being attended again. IoR mechanisms have been widely used in visual attention models (see \cite{itti2001computational} for a review) to suppress the attended location and avoid endless attractions towards the most salient stimulus.

\subsection{State}
\label{subsec:state}

The state representation is a tuple $(o,h)$, where $o$ is a feature vector of the observed region, and $h$ is a vector with the history of taken actions. The set of possible states $S$ is very large as it includes arbitrary boxes from a large set of images, and is expanded with all the combinations of actions that lead to those boxes. Therefore, generalization is important to design an effective state representation.

The feature vector $o$ is extracted from the current region using a pre-trained CNN following the architecture of Zeiler and Fergus \cite{zeiler2014visualizing}. Any attended region by the agent is warped to match the input of the network ($224\times224$) regardless of its size and aspect ratio, following the technique proposed by Girshick et al. \cite{girshick2014rich}. We also expand the region to include 16 pixels of context around the original box. We forward the region up to the layer 6 (fc6) and use the 4,096 dimensional feature vector to represent its content.

The history vector $h$ is a binary vector that informs which actions have been used in the past. Each action in the history vector is represented by a 9-dimensional binary vector, where all values are zero except the one corresponding to the taken action. The history vector encodes 10 past actions, which means that $h\in\mathbb{R}^{90}$. Although $h$ is very low-dimensional compared to $o$, it has enough energy to inform what has happened in the past. This information demonstrated to be useful to stabilize search trajectories that might get stuck in repetitive cycles, improving average precision by approximately 3 percent points. The history vector also works better than appending a few more frames to the state representation with the additional benefit of increasing dimensionality by a negligible factor.

\subsection{Reward Function}
\label{subsec:reward}

The reward function $R$ is proportional to the improvement that the agent makes to localize an object after selecting a particular action. Improvement in our setup is measured using the Intersection-over-Union (IoU) between the target object and the predicted box at any given time. More specifically, the reward function is estimated using the differential of IoU from one state to another. The reward function can be estimated during the training phase only because it requires ground truth boxes to be calculated. 

Let $b$ be the box of an observable region, and $g$ the ground truth box for a target object. Then, IoU between $b$ and $g$ is defined as $IoU(b,g) = {area( b \cap g )}/{area( b \cup g )}$.

The reward function $R_a(s,s')$ is granted to the agent when it chooses the action $a$ to move from state $s$ to $s'$. Each state $s$ has an associated box $b$ that contains the attended region. Then, the reward is as follows\footnote{Notice that the ground truth box $g$ is part of the environment and cannot be modified by the agent.}:

\begin{equation} \label{eq:reward}
R_a(s,s') = sign\left( IoU(b',g) - IoU(b,g) \right)
\end{equation}

Intuitively, equation \ref{eq:reward} says that the reward is positive if IoU improved from state $s$ to state $s'$, and negative otherwise. This reward scheme is binary $r\in\{-1, +1\}$, and applies to any action that transforms the box. Without quantization, the difference in IoU is small enough to confuse the agent about which actions are good or bad choices. Binary rewards communicate more clearly which transformations keep the object inside the box, and which ones take the box away the target. In this way, the agent pays a penalty for taking the box away the target, and is rewarded to keep the target object in the visible region until there is no other transformation that improves localization. In that case, the best action to choose should be the trigger.

The trigger has a different reward scheme because it leads to a terminal state that does not change the box, and thus, the differential of IoU will always be zero for this action. The reward for the trigger is a thresholding function of IoU as follows:

\begin{equation} \label{eq:reward2}
R_\omega(s,s') =
\left\{
	\begin{array}{ll}
		+\eta  & \mbox{if } IoU(b,g) \geq \tau \\
		-\eta & \mbox{otherwise }
	\end{array}
\right.
\end{equation}

where $\omega$ is the trigger action, $\eta$ is the trigger reward, set to $3.0$ in our experiments, and $\tau$ is a threshold that indicates the minimum IoU allowed to consider the attended region as a positive detection. The standard threshold for object detection evaluation is 0.5, but we used $\tau=0.6$ during training to encourage better localization. A larger value for $\tau$ has a negative effect in performance because the agent learns that only clearly visible objects are worth the trigger, and tends to avoid truncated or naturally occluded objects.

Finally, the proposed reward scheme implicitly considers the number of steps as a cost because of the way in which Q-learning models the discount of future rewards (positive and negative). The agent follows a greedy strategy, which prefers short sequences because any unnecessary step pays a penalty that reduces the accumulated utility.

\section{Finding a Localization Policy with Reinforcement Learning}
\label{sec:solution}

The goal of the agent is to transform a bounding box by selecting actions in a way that maximizes the sum of the rewards received during an interaction with the environment (an episode). The core problem is to find a policy that guides the decision making process of this agent. A policy is a function $\pi(s)$ that specifies the action $a$ to be chosen when the current state is $s$. Since we do not have the state transition probabilities and the reward function is data-dependent, the problem is formulated as a reinforcement learning problem using Q-learning \cite{sutton1998reinforcement}.

In our work, we follow the deep Q-learning algorithm recently proposed by Mnih et al. \cite{mnih2015human}. This approach estimates the action-value function using a neural network, and has several advantages over previous Q-learning methods. First, the output of the Q-network has as many units as actions in the problem. This makes the model efficient because the input image is forwarded through the network only once to estimate the value of all possible actions. Second, the algorithm incorporates a replay-memory to collect various experiences and learns from them in the long run. In this way, transitions in the replay-memory are used in many model updates resulting in greater data efficiency. Third, to update the model the algorithm selects transitions from the replay-memory uniformly at random to break short-term correlations between states. This makes the algorithm more stable and prevents divergence of the parameters. After learning the action-value function $Q(s,a)$, the policy that the agent follows is to select the action $a$ with the maximum estimated value.

\subsection{Q-learning for Object Localization}

\begin{figure}[t]
\begin{center}
  \includegraphics[width=0.9\linewidth]{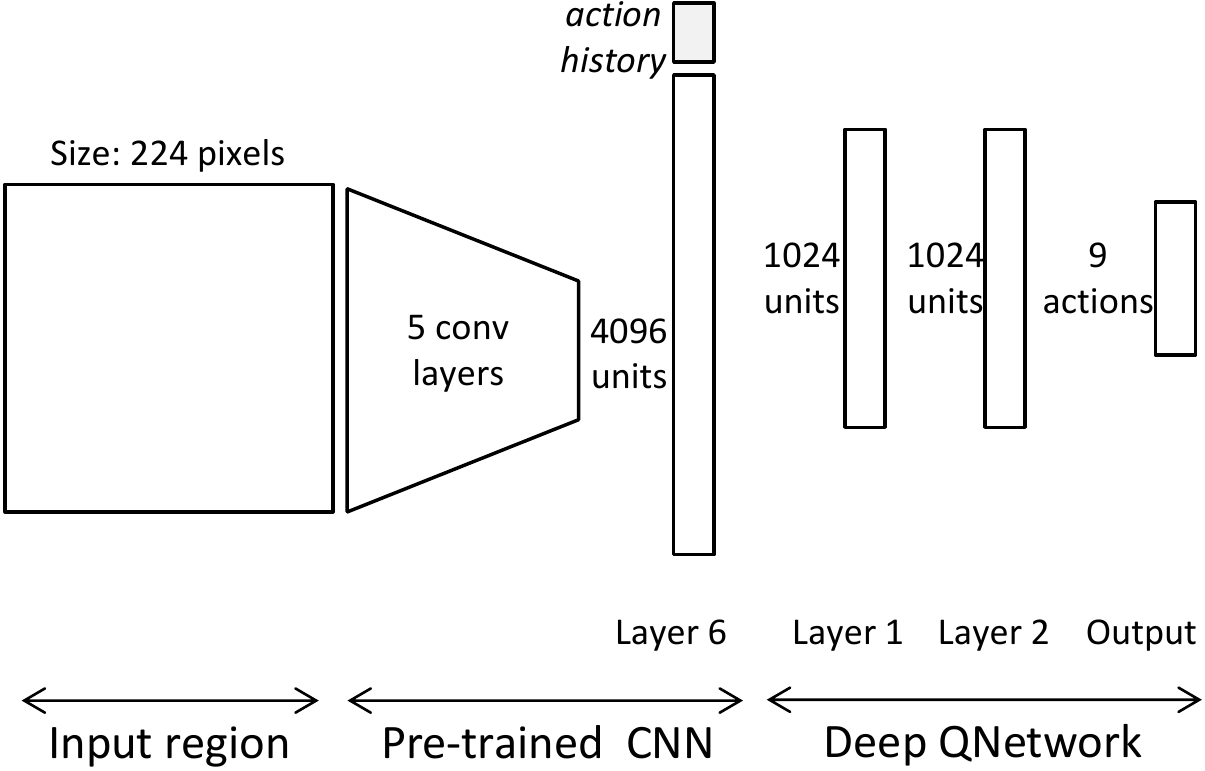}
\end{center}
   \caption{Architecture of the proposed QNetwork. The input region is first warped to $224\times224$ pixels and processed by a pre-trained CNN with 5 convolutional layers and 1 fully connected layer. The output of the CNN is concatenated with the action history vector to complete the state representation. It is processed by the Q-network which predicts the value of the 9 actions.}
\label{fig:qnet}
\end{figure}

We use a Q-network that takes as input the state representation discussed in section \ref{subsec:state} and gives as output the value of the nine actions presented in section \ref{subsec:actions}. We train category specific Q-networks following the architecture illustrated in Figure \ref{fig:qnet}. Notice that in our design we do not learn the full feature hierarchy of the convolutional network; instead, we rely on a pre-trained CNN.

Using a pre-trained CNN has two advantages: First, learning the Q function is faster because we need to update the parameters of the Q-Network only, while using the deep CNN just as a feed-forward feature extractor. Second, the hierarchy of features is trained with a larger dataset, leveraging generic discriminative features in our method. Learning the full hierarchy of features is possible under the deep Q-learning framework, and we hypothesize that performance could be improved because the learned features would be adapted for the localization task instead of classification. However, this mainly requires larger object detection datasets, so we leave this possibility for future work.

\subsection{Training Localization Agents}
\label{subsec:training}

The parameters of the Q-network are initialized at random. Then, the agent is set to interact with the environment in multiple episodes, each presenting a different training image. The policy followed during training is $\epsilon$-greedy \cite{sutton1998reinforcement}, which gradually shifts from exploration to exploitation according to the value of $\epsilon$. During exploration, the agent selects actions randomly to observe different transitions and collects a varied set of experiences. During exploitation, the agent selects actions greedily according to the learned policy, and learns from its own successes and mistakes.

In our work, exploration does not proceed with random actions. Instead, we use a guided exploration strategy following the principles of apprenticeship learning \cite{abbeel2004apprenticeship, coates2008learning, levine2015endtoend}, which is based on demonstrations made by an expert to the agent. Since the environment knows the ground truth boxes, and the reward function is calculated with respect to the IoU with the current box, we can identify which actions will give positive and negative rewards. Then, during exploration, the agent is allowed to choose one random action from the set of positive actions \footnote{Or any action if this set of positive actions is empty}. Notice that for one particular state $s$, there might be multiple positive actions because there is no single path to localize an object. Using this strategy, the algorithm terminates in a small number of epochs.

The $\epsilon$-greedy training strategy is run for 15 epochs, each completed after the agent has had interaction with all training images. During the first 5 epochs, $\epsilon$ is annealed linearly from 1.0 to 0.1 to progressively let the agent use its own learned model. After the fifth epoch, $\epsilon$ is fixed to 0.1 so the agent adjusts the model parameters from experiences produced by its own decisions. The parameters are updated using stochastic gradient descent and the back-propagation algorithm, and we also use dropout regularization \cite{srivastava2014dropout}.

\subsection{Testing a Localization Agent}
\label{subsec:testing}

Once an agent is trained with the procedure described above, it learns to attend regions that contain objects of the target category. Since we do not know the number of objects present in a single image beforehand, we let the agent run for a maximum of 200 steps, so only 200 regions are evaluated per test image. An alternative to stop the search after a fixed number of steps is to include an extra termination action to let the agent indicate when the search is done. However, additional actions introduce new errors and make the problem more difficult to learn. We simplified the model with a minimum set of actions to act locally in time.

At each step, the agent makes a decision to transform the current box or selects the trigger to indicate that an object has been found. When the trigger is used, the search for other objects continues from a new box that covers a large portion of the image. The search for objects is restarted from the beginning due to two possible events: the agent used the trigger, or 40 steps passed without using the trigger. We found that 40 steps are enough to localize most objects, even the smaller ones, and when it takes longer is usually because the agent is stuck searching in an ambiguous region. Restarting the box helps to take a new perspective of the scene. The very first box covers the entire scene, and after any restarting event, the new box has a reduced size set to 75\% of the image size, and is placed in one of the four corners of the image always in the same order (from top-left to right-bottom).

\section{Experiments and Results}
\label{sec:experiments}

We evaluated the performance of the proposed model using the Pascal VOC dataset. The localization method is sensitive to the amount of data used for training, and for that reason, we used the combined training sets of 2007 and 2012. Using this joint set, results improved nearly 10\% relative to using either one alone. We evaluate performance on the test set of VOC 2007 and report our findings under this setting.

\subsection{Evaluation of Precision}
\label{subsec:precision}

As an object detector, our algorithm can be evaluated in two modes: 1) All attended regions (AAR), a detector that scores all regions processed by the agent during a search episode. This is useful to consider well-localized regions that were not explicitly marked by the agent as detections. 2) Terminal regions (TR), a detector that only considers regions in which the agent explicitly used the trigger to indicate the presence of an object. In both cases, we use an external linear SVM trained with the same procedure as R-CNN (with hard-negative mining on region proposals using the VOC2012 training set) to score the attended regions. This classifier is useful to rerank candidate regions assuming that our model generates object proposals. The scores computed by the Q-network are not useful for object detection evaluation because they estimate the value of actions instead of discriminative scores.

\begin{table*}
\begin{center}
{\scriptsize
\begin{tabular}{l|p{0.3cm}p{0.3cm}p{0.3cm}p{0.3cm}p{0.3cm}p{0.3cm}p{0.3cm}p{0.3cm}p{0.3cm}p{0.3cm}p{0.3cm}p{0.3cm}p{0.3cm}p{0.4cm}p{0.4cm}p{0.3cm}p{0.3cm}p{0.3cm}p{0.3cm}p{0.4cm}|c}
Method&aero&bike&bird&boat&bottle&bus&car&cat&chair&cow&table&dog&horse&mbike&person&plant&sheep&sofa&train&tv&MAP\\
\hline
DPM \cite{felzenszwalb2010object}&33.2&60.3&10.2&16.1&\textbf{27.3}&54.3&58.2&23.0&20.0&24.1&26.7&12.7&58.1&48.2&43.2&12.0&21.1&36.1&46.0&43.5&33.7\\
\hline
MultiBox \cite{erhan2014scalable}&41.3&27.7&30.5&17.6&3.2&45.4&36.2&53.5&6.9&25.6&27.3&46.4&31.2&29.7&37.5&7.4&29.8&21.1&43.6&22.5&29.2\\
DetNet \cite{szegedy2013deep}&29.2&35.2&19.4&16.7&3.7&53.2&50.2&27.2&10.2&34.8&30.2&28.2&46.6&41.7&26.2&10.3&32.8&26.8&39.8&47.0&30.5\\
Regionlets \cite{zou2014generic}&44.6&55.6&24.7&23.5&6.3&49.4&51.0&\textbf{57.5}&14.3&35.9&45.9&41.3&\emph{\underline{61.9}}&54.7&44.1&16.0&28.6&\textbf{41.7}&\emph{\underline{63.2}}&44.2&40.2\\
\hline
Ours TR&\textbf{57.9}&56.7&38.4&33.0&17.5&51.1&52.7&53.0&17.8&39.1&\emph{\underline{47.1}}&52.2&58.0&\textbf{57.0}&45.2&19.3&42.2&35.5&54.8&49.0&43.9\\
Ours AAR&55.5&\textbf{61.9}&\textbf{38.4}&\textbf{36.5}&21.4&\textbf{56.5}&\textbf{58.8}&55.9&\textbf{21.4}&\textbf{40.4}&46.3&\textbf{54.2}&56.9&55.9&\textbf{45.7}&\textbf{21.1}&\textbf{47.1}&41.5&54.7&\textbf{51.4}&\textbf{46.1}\\
\hline
R-CNN \cite{girshick2014rich}&\emph{\underline{64.2}}&\emph{\underline{69.7}}&\emph{\underline{50.0}}&\emph{\underline{41.9}}&\emph{\underline{32.0}}&\emph{\underline{62.6}}&\emph{\underline{71.0}}&\emph{\underline{60.7}}&\emph{\underline{32.7}}&\emph{\underline{58.5}}&\textbf{46.5}&\emph{\underline{56.1}}&\textbf{60.6}&\emph{\underline{66.8}}&\emph{\underline{54.2}}&\emph{\underline{31.5}}&\emph{\underline{52.8}}&\emph{\underline{48.9}}&\textbf{57.9}&\emph{\underline{64.7}}&\emph{\underline{54.2}}\\
\end{tabular}
}
\end{center}
\caption{Average Precision (AP) per category in the Pascal VOC 2007 test set. The DPM system is the only baseline that does not use CNN features. R-CNN is the only method that uses object proposals. Our system is significantly better at localizing objects than other recent systems that predict bounding boxes from CNN features without object proposals. Numbers in bold are the second best result per column, and underlined numbers are the overall best result.}
\label{tab:ap2007}
\end{table*}

We compare performance with other methods recently proposed in the literature. Table \ref{tab:ap2007} presents detailed, per-category Average Precision (AP) for all systems. The only baseline method that does not use CNN features is the DPM system \cite{felzenszwalb2010object}. MultiBox \cite{erhan2014scalable} and DetNet \cite{szegedy2013deep} predict bounding boxes from input images using deep CNNs with a regression objective, and the work of Zou et al. \cite{zou2014generic} adapts the regionlets framework with CNN features too. Finally, we compare with the R-CNN system \cite{girshick2014rich} configured with a network architecture that has the same number of layers as ours, and without bounding box regression.

Overall, the R-CNN system still has the best performance and remains as a strong baseline. The major drawback of R-CNN is that it relies on a large number of object proposals to reach that performance, and demands significant computing power. MultiBox and Regionlets attempt to leverage deep CNNs in more efficient ways, by using only a few boxes to make predictions or avoiding re-computing features multiple times. Our approach also aims to localize objects by attending to a small number of regions, and this has an impact in performance. However, our result is significantly better than the other baseline methods, reaching 46.1 MAP, while the next best reaches 40.2 MAP.

The main difference in performance between the TR and the ARR settings is that the first one only ranks regions explicitly marked by the agent with the trigger action. The difference in performance is 2.2 percent points, which is relatively small, considering that TR proposes an average of only 1.3 regions per category per image (including true and false positives in all test images). Both settings require the same computational effort\footnote{The cost of the classifier is negligible compared to the cost of feature extraction and analysis.}, because the agent has to attend the same number of regions. However, TR is able to identify which of those regions are promising objects with very high accuracy.

\subsection{Evaluation of Recall}
\label{subsec:recall}

\begin{figure}[t]
\begin{center}
  \begin{overpic}[scale=0.45]{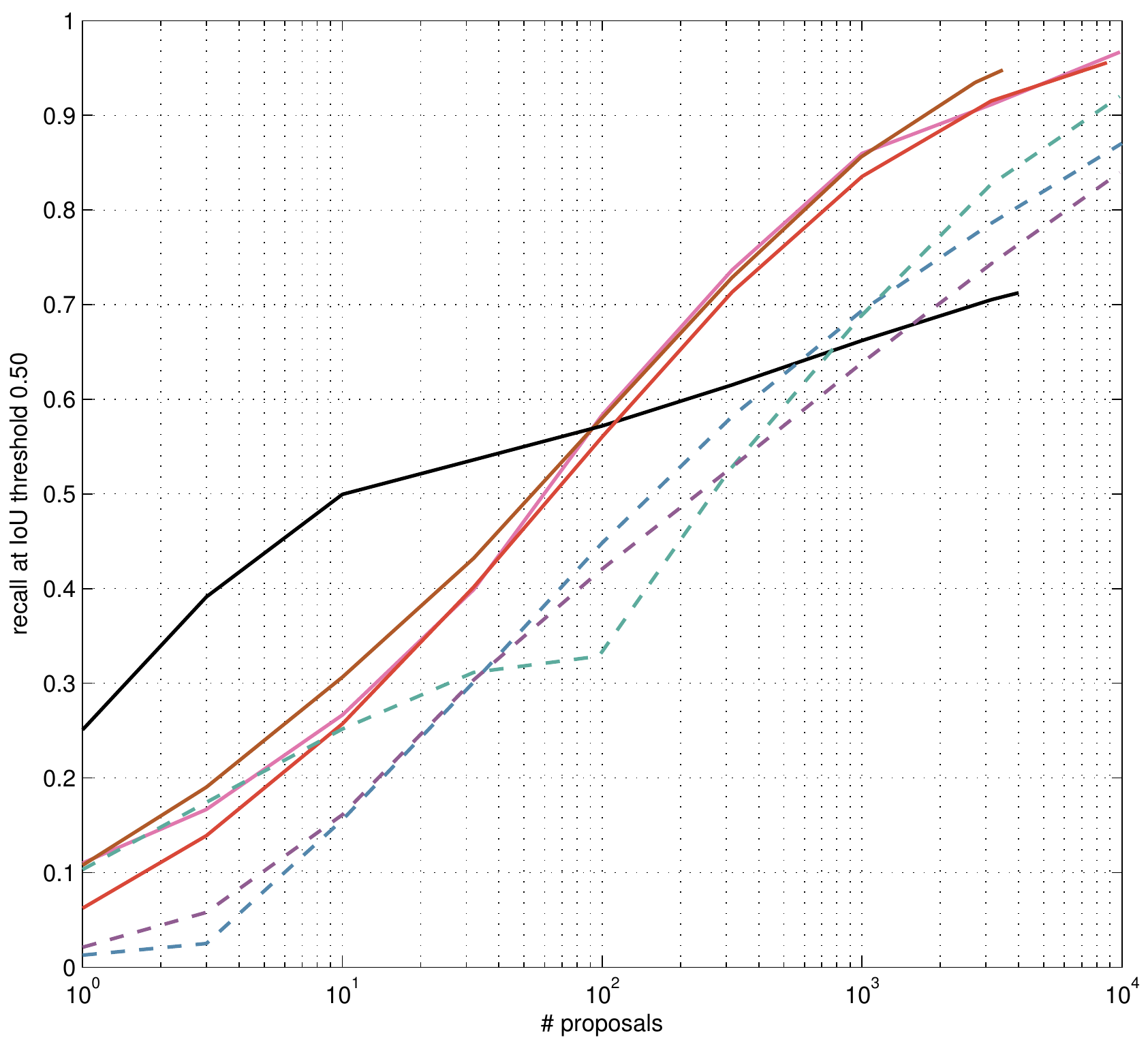}
	\put(10,65){\includegraphics[width=0.2\linewidth]{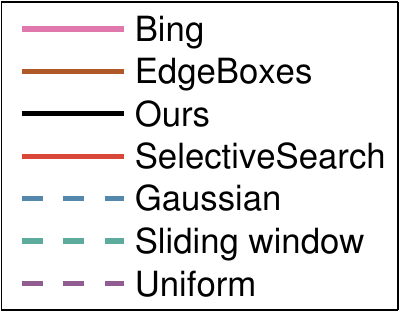}}
	\end{overpic}
\end{center}
   \caption{Recall as a function of the number of proposed regions. Solid lines are state-of-the-art methods, and dashed lines are simple baselines. Our approach is significantly better at early recall: only 10 proposals per image reach 50\% recall. The tendency is also fundamentally different: overall recall does not depend strongly on a large number of proposed regions. Notice that our approach is not a category-independent region proposal algorithm.}
\label{fig:recall}
\end{figure}

All the regions attended by the agent can be understood as object proposal candidates, so we evaluate them following the methodology proposed by Hosang et al. \cite{hosang2014good}. Overall, our method running for 200 steps per category, processes a total of 4,000 candidates per image reaching 71\% of recall. This is between 10 and 25 points less recall than most methods achieve at a similar number of proposals. However, the recall of our proposals is significantly superior for the top 100 candidates per image. 

For this evaluation we score attended regions using the Q-values predicted by the agent, and we add a large constant only to those regions for which the agent used the trigger. This gives priority to regions that the agent considers correctly localized objects. We use this scoring function instead of the classification scores for fairness in the evaluation \cite{chavali2015object}, since other methods rank proposals using an estimated objectness score, and high Q-values can be interpreted in the same way. Figure \ref{fig:recall} shows the number of proposals vs. recall for several baseline proposal methods. In this evaluation, we include selective search \cite{uijlings2013selective}, BING \cite{cheng2014bing} and EdgeBoxes \cite{zitnick2014edge}. The results show that our method reaches 50\% of recall with just 10 proposals per image, while all the other methods are around or below 30\%.

\begin{figure}[t]
\begin{center}
  \includegraphics[width=1.0\linewidth]{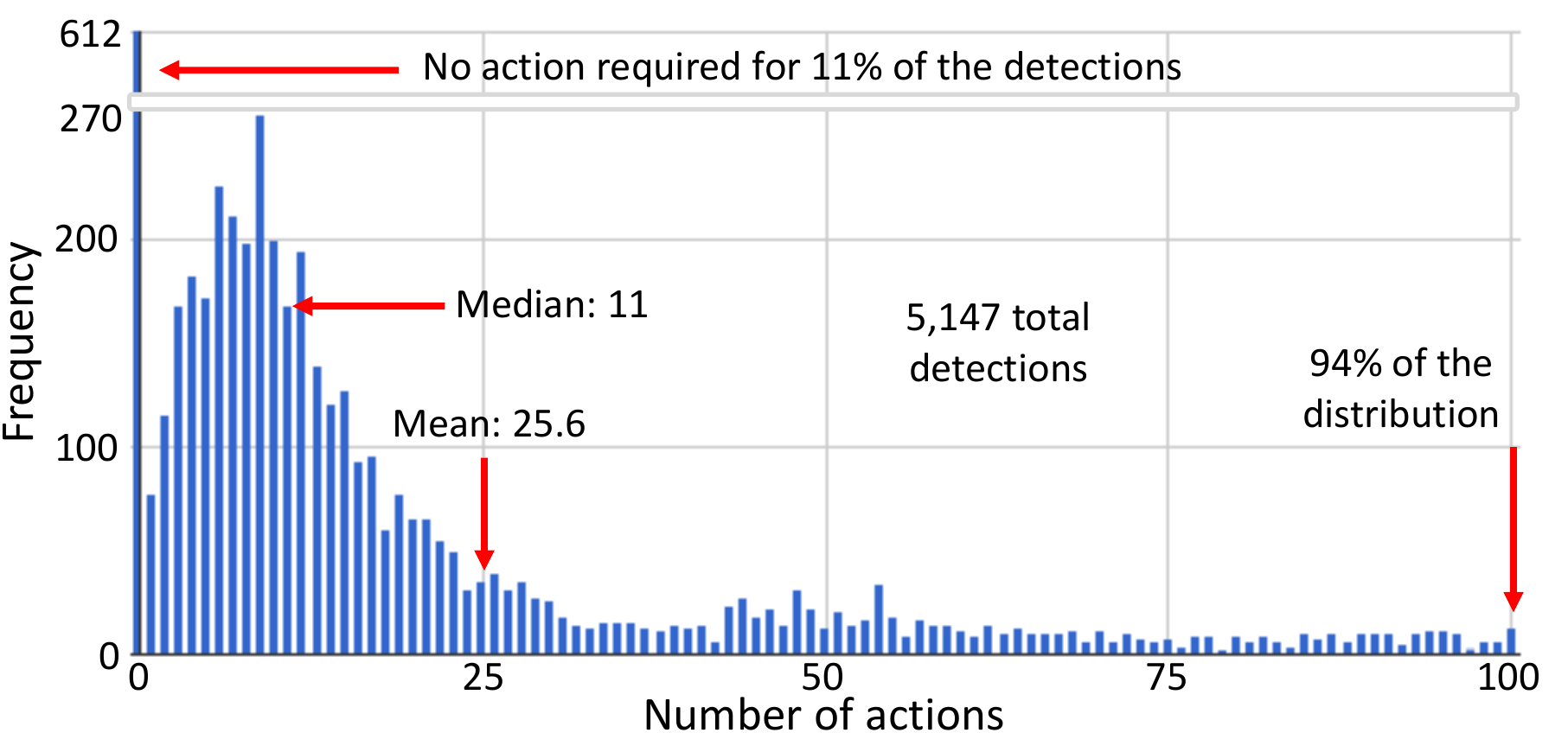}
\end{center}
   \caption{Distribution of detections explicitly marked by the agent as a function of the number of actions required to reach the object. For each action, one region in the image needs to be processed. Most detections can be obtained with about 11 actions only.}
\label{fig:distribution}
\end{figure}

This evaluation emphasizes two important points: first, our method uses category-specific knowledge to find objects, and this is clearly an advantage over category-independent proposals, but is also a drawback since objects that cannot be recognized given the feature representation are never localized. Second, the main challenge of our approach is to improve overall recall, that is, to localize every single object without missing any instance. Notice that more candidates do not improve recall as much as other methods do, so we hypothesize that fixing overall recall will improve early recall even more.

To illustrate this result further, we plot the distribution of correctly detected objects according to the number of steps necessary to localize them in Figure \ref{fig:distribution}. The distribution has a long tail, with 83\% of detections requiring less than 50 steps to be obtained, and an average of 25.6. A more robust statistic for long-tailed distributions is the median, which in our experiments is just 11 steps, indicating that most of the correct detections happen around that number of steps. Also, the agent is able to localize 11\% of the objects immediately without processing more regions, because they are big instances that occupy most of the image.

\subsection{Qualitative Evaluation}
\label{subsec:qualitative}

\begin{figure}[t]
\begin{center}
	\includegraphics[width=1.0\linewidth]{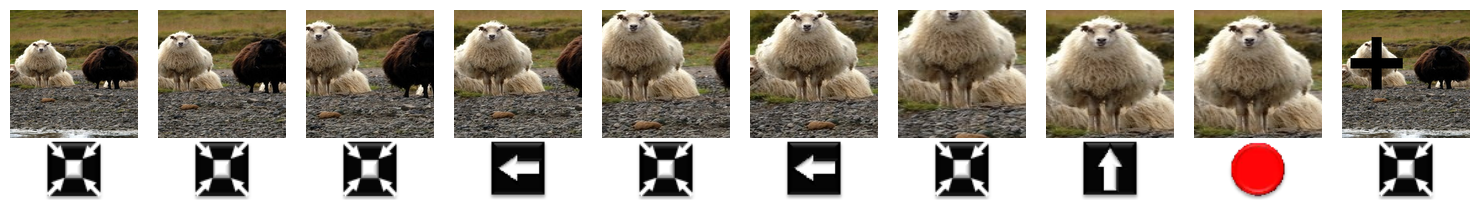}\\
	\includegraphics[width=1.0\linewidth]{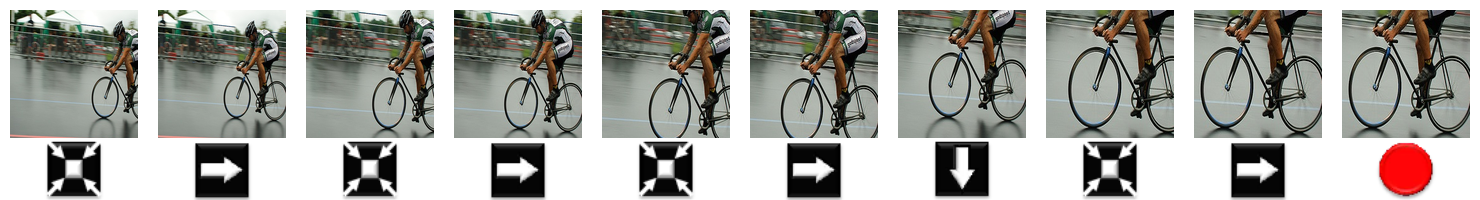}\\
	\includegraphics[width=1.0\linewidth]{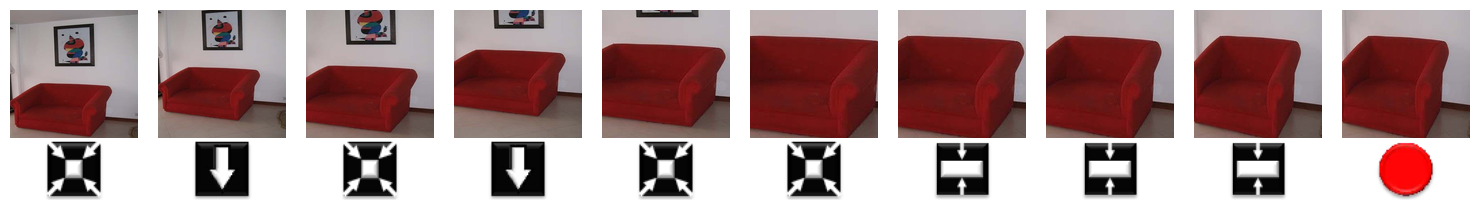}\\
	\includegraphics[width=1.0\linewidth]{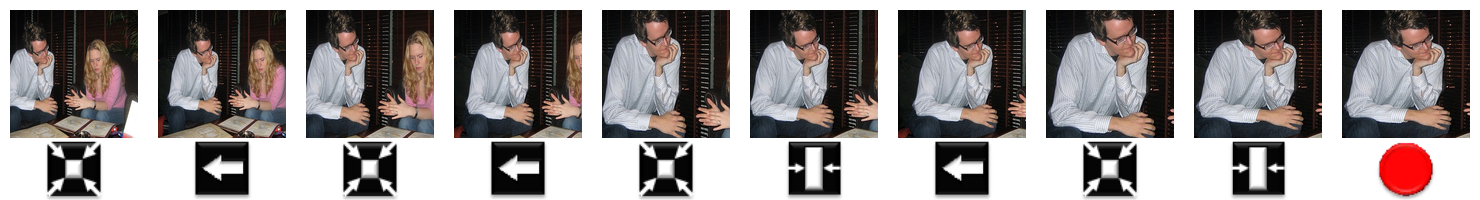}\\
	\includegraphics[width=1.0\linewidth]{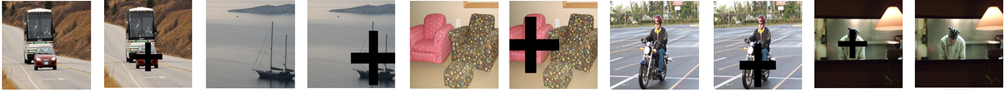}
\end{center}
   \caption{Example sequences observed by the agent and the actions selected to focus objects. Regions are warped in the same way as they are fed to the CNN. Actions keep the object in the center of the box. More examples in the supplementary material. Last row: example Inhibition of Return marks placed during test.}
\label{fig:sequences}
\end{figure}

\begin{figure}[t]
\begin{center}
  \includegraphics[width=1.00\linewidth]{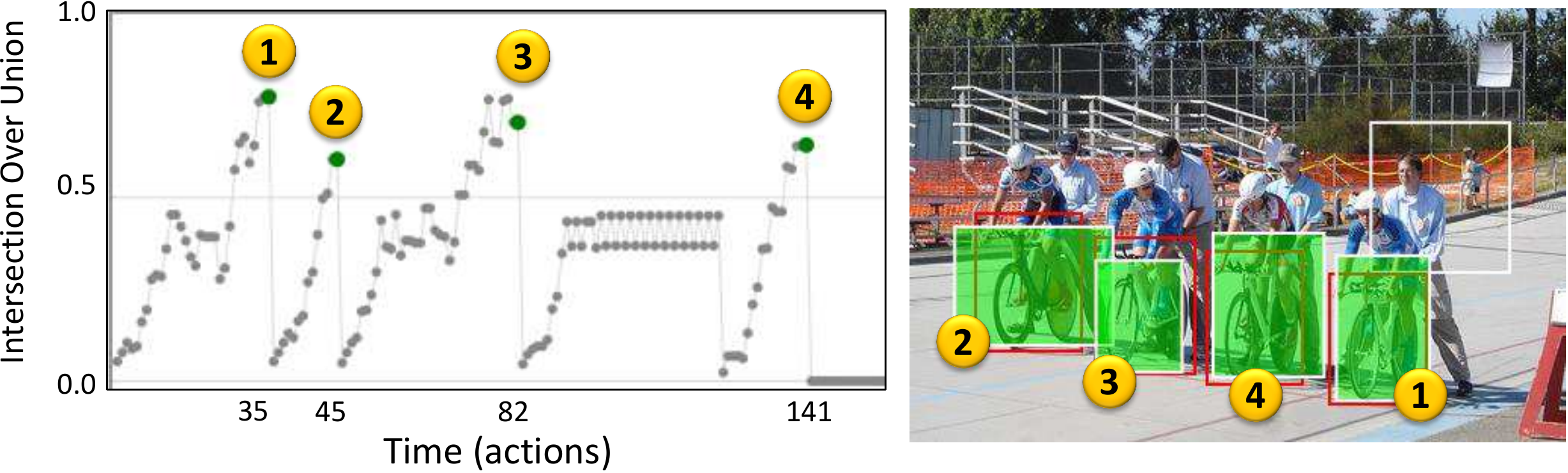}\\
	\includegraphics[width=1.00\linewidth]{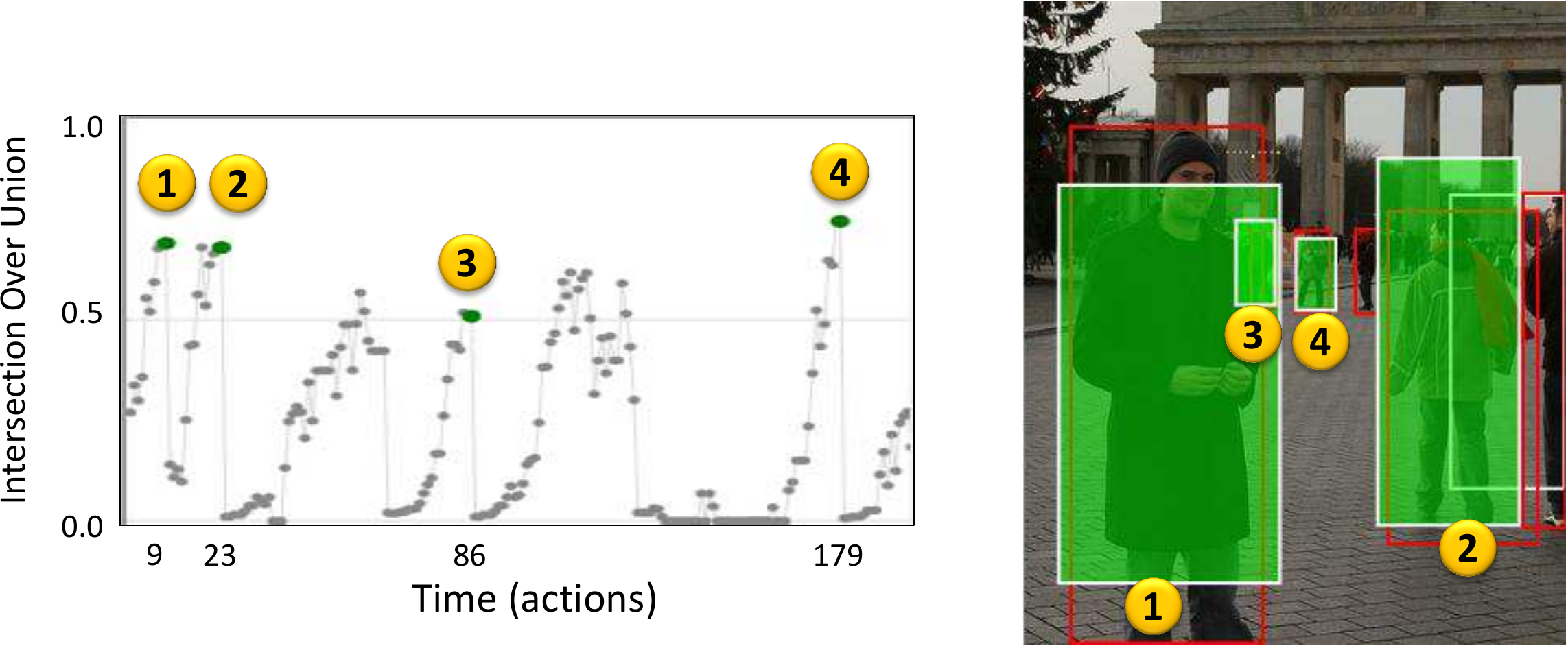}
\end{center}
   \caption{Examples of multiple objects localized by the agent in a single scene. Numbers in yellow indicate the order in which each instance was localized. Notice that IoU between the attended region and ground truth increases quickly before the trigger is used.}
\label{fig:examples}
\end{figure}

We present a number of example sequences of regions that the agent attended to localize objects. Figure \ref{fig:examples} shows two example scenes with multiple objects, and presents green boxes where a correct detection was explicitly marked by the agent. The plot to the left presents the evolution of IoU as the agent transforms the bounding box. These plots show that correct detections are usually obtained with a small number of steps increasing IoU with the ground truth rapidly. Points of the plots that oscillate below the minimum accepted threshold (0.5) indicate periods of the search process that were difficult or confusing to the agent.

Figure \ref{fig:sequences} shows sequences of attended regions as seen by the agent, as well as the actions selected in each step. Notice that the actions chosen attempt to keep the object in the center of the box, and also that the final object appears to have normalized scale and aspect ratio. The top two examples also show the IoR mark that is placed in the environment after the agent triggers a detection. The reader can find more examples and videos in the supplementary material.

\subsection{Error Modes}
\label{subsec:errors}

\begin{figure}[t]
\begin{center}
  \includegraphics[width=1.00\linewidth]{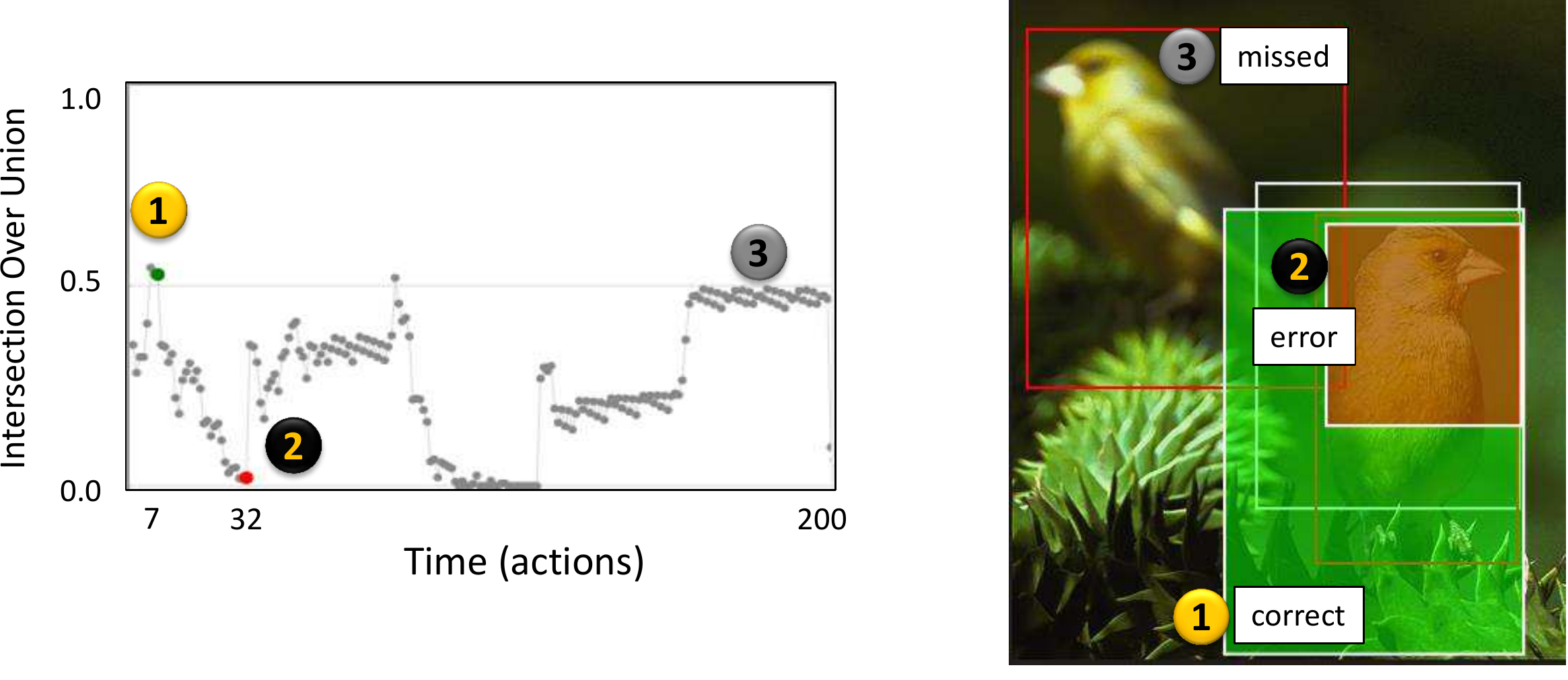}\\
	\includegraphics[width=1.00\linewidth]{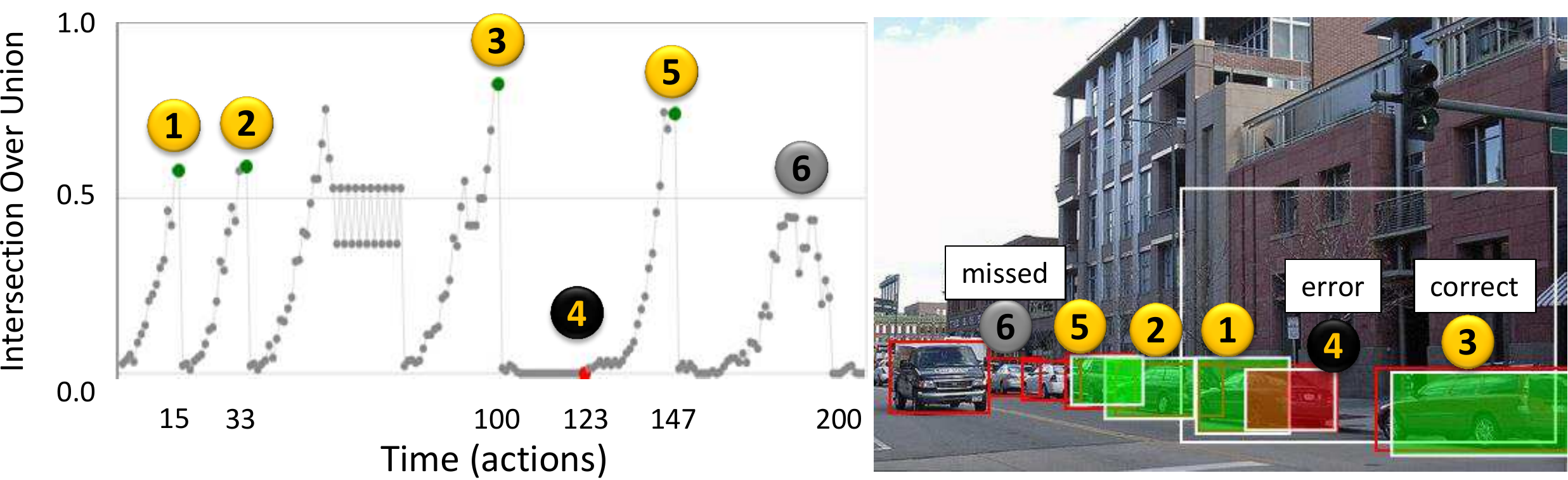}
\end{center}
   \caption{Examples of images with common mistakes that include: duplicated detections due to objects not fully covered by the IoR mark, and missed objects due to size or other difficult patterns.}
\label{fig:errors}
\end{figure}

\begin{figure}[t]
\begin{center}
	\includegraphics[width=0.48\linewidth]{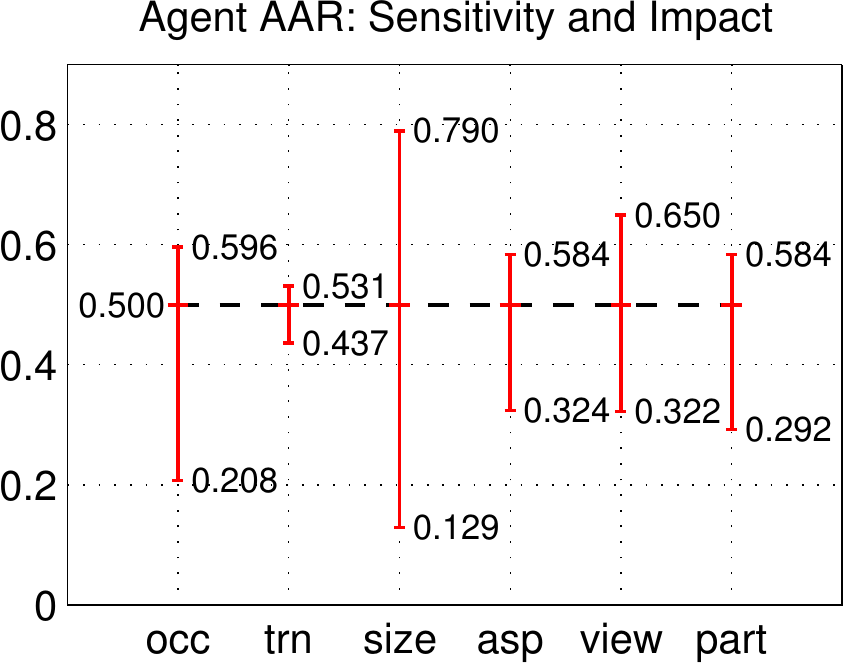}
	\includegraphics[width=0.48\linewidth]{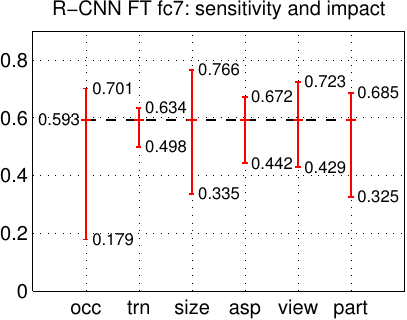}
\end{center}
   \caption{Sensitivity analysis following the method of Hoiem et al. \cite{hoiem2012diagnosing}. Error bars are the average normalized AP over categories with the highest and lowest performance for each characteristic. Evaluated characteristics are: occlusion (occ), truncation (trn), size, aspect ratio (asp), viewpoint of objects (view), and visibility of parts (parts).}
\label{fig:sensitivity}
\end{figure}

We also evaluate performance using the diagnostic tool proposed by Hoiem et al. \cite{hoiem2012diagnosing}. In summary, object localization is the most frequent error of our system, and it is sensitive to object size. Here we include the report of sensitivity to characteristics of objects in Figure \ref{fig:sensitivity}, and compare to the R-CNN system. Our system is more sensitive to the size of objects than any other characteristic, which is mainly explained by the difficulty of the agent to attend cluttered regions with small objects. Interestingly, our system seems to be less sensitive to occlusion and truncation than R-CNN.

Figure \ref{fig:errors} presents object search episodes with some common errors. The figure shows examples of objects that the agent tried to localize but IoU never passed the minimum threshold. Those objects are missed because none of the attended regions surpass the threshold. In both examples, the agent triggered an additional detection in a location where a correct detection was placed before. This happens because the inhibition-of-return (IoR) mark leaves part of the object visible that may be observed again by the agent. 

The IoR mark also generates other types of errors such as covering additional objects in the scene that cannot be recovered. We found that 78\% of all missed detections happen in images with multiple instances of the same object, and the other 22\% occur in images with a single instance. Not all missed objects are explained by obstruction of the IoR mark, however, we experimented with different strategies to change the focus of attention. We used a black box of the same size of the predicted box, but this results in a highly intrusive mark. We also experimented with a memory of detected instances to give negative rewards for going back to a known region, but training becomes unstable. The final cross that we adopted is centered in the predicted box and covers a third of its area, leaving parts of the region visible in favor of other overlapping objects. Overlapping objects are in general a challenge for most object detectors, and the problem requires special research attention.

\subsection{Runtime}
\label{subsec:runtime}

We conducted runtime experiments using a workstation equipped with a K-40 GPU. Feature extraction and action-decision run in the GPU. Our algorithm proceeds by analyzing a single region at a time, and processing each region takes an average of 7.74ms. This time is divided between feature extraction with the CNN (4.5 ms) and decision making with the Q-network (3.2ms), imposing an overhead of about 70\% more computing power per region in our prototype, compared to systems that only extract features. This overhead is likely to be reduced using a more optimized implementation. Since we run the system for 200 steps, the total test processing time is 1.54s in average (wall clock time). Notice that we do not add segmentation or any other pre-processing time, because our system directly decides which boxes need to be evaluated.

\section{Conclusions and Future Work}
\label{sec:conclusions}

A system that learns to localize objects in scenes using an attention-action strategy has been presented. This approach is fundamentally different from previous works for object detection, because it follows a top-down scene analysis to narrow down the correct location of objects. Reinforcement learning (RL) demonstrated to be an efficient strategy to learn a localization policy, which is a challenging task since an object can be localized following different search paths. Nevertheless, the proposed agent learns from its own mistakes and optimizes the policy to find objects.

Experiments show that the system can localize a single instance of an object processing between 11 and 25 regions only. This results in a very efficient strategy for applications where a few number of categories are required. Our formulation needs to be extended in other ways for scaling to large numbers of categories; for instance, making it category-independent or using hierarchical ontologies to decide the fine-grained category later. An important challenge to improve recall needs to be addressed. Part of our future work includes training the system end-to-end instead of using a pre-trained CNN, and using deeper CNN architectures for improving the accuracy of predictions.

\textbf{Acknowledgements}. Special thanks to David Forsyth and Derek Hoiem for discussions and advice, and to Oscar S\'anchez for his feedback. We gratefully acknowledge NVIDIA corporation for the donation of Tesla GPUs for this research. This research was part of the Blue Waters sustained-petascale computing project. This work was partially supported by NSF grants IIS 1228082, CIF 1302438, and the DARPA Computer Science Study Group (D12AP00305).

{\small
\bibliographystyle{ieee}
\bibliography{locAgents}
}

\end{document}